\documentclass{article}

\usepackage{PRIMEarxiv}

\usepackage[utf8]{inputenc} 
\usepackage[T1]{fontenc}    
\usepackage{hyperref}       
\hypersetup{colorlinks,allcolors=blue}
\usepackage{url}            
\usepackage{booktabs}       
\usepackage{amsfonts}       
\usepackage{nicefrac}       
\usepackage{microtype}      
\usepackage{lipsum}
\usepackage[letterpaper]{geometry}
\usepackage{amsmath,amssymb,mathtools}
\usepackage{bbm}
\usepackage{hyperref}
\usepackage{xcolor}
\usepackage{multirow}
\usepackage{array, makecell}
\usepackage{booktabs}
\usepackage{fancyhdr}       
\usepackage{graphicx}       
\usepackage{amsthm}

\graphicspath{{media/}}     

\newtheorem{theorem}{Theorem}
\newtheorem{Definition}{Definition}

\title{NSOTree: Neural Survival Oblique Tree
}

\author{
  Xiaotong Sun \\
  Department of Industrial Engineering \\
  University of Arkansas \\
  Fayetteville\\
  \texttt{xs018@uark.edu} \\
   \And
  Peijie Qiu \\
  McKeley School of Engineering \\
  Washington University in St. Louis\\
  St. Louis\\
  \texttt{peijie.qiu@wustl.edu} \\
}

\begin{document}
\maketitle

\begin{abstract}
Survival analysis is a statistical method employed to scrutinize the duration until a specific event of interest transpires, known as time-to-event information characterized by censorship. Recently, deep learning-based methods have dominated this field due to their representational capacity and state-of-the-art performance. However, the black-box nature of the deep neural network hinders its interpretability, which is desired in real-world survival applications but has been largely neglected by previous works. In contrast, conventional tree-based methods are advantageous with respect to interpretability, while consistently grappling with an inability to approximate the global optima due to greedy expansion. In this paper, we leverage the strengths of both neural networks and tree-based methods, capitalizing on their ability to approximate intricate functions while maintaining interpretability. To this end, we propose a Neural Survival Oblique Tree (NSOTree) for survival analysis. Specifically, the NSOTree was derived from the ReLU network and can be easily incorporated into existing survival models in a plug-and-play fashion. Evaluations on both simulated and real survival datasets demonstrated the effectiveness of the proposed method in terms of performance and interpretability.
\end{abstract}

\keywords{Survival Analysis \and Deep Neural Networks \and Oblique Tree \and Tree-based method}

\section{Introduction}
In the realm of statistics, survival analysis is a prominent area of interest, employed for modeling time-to-event data across multifarious fields including clinical research, economics, and engineering. Time-to-event data exhibits to be censored due to the lack of follow-up records, the termination of study before all events occur, etc. This form of censorship can manifest as right-censored, left-censored, or interval-censored, contingent upon whether the event time falls after, before, or within the observation period~\cite{lee2003statistical}. The censored data poses significant challenges in survival analysis because the true distribution of the event time is unknown. 

In the early stages, survival analysis is commonly approached from the perspective of traditional statistics. Within this context, the non-parametric Kaplan-Meier estimator~\cite{kaplan1958nonparametric} is employed to address censored data by applying counting statistics to event time. However, it does not take the time-dependent covariates into consideration. The Cox proportional hazards (CPH) model~\cite{cox1992regression} addresses this limitation by proposing a semi-parametric model that approximates the hazard rate function by two components: i) a baseline hazard rate and ii) a linear combination of covariates. Nevertheless, this linear relationship between the risk function and covariates can be easily violated in real-world applications. 
Despite extensive efforts, the limited capabilities of traditional statistical survival models in capturing complex event-time distribution make them hard to generalize well to modern personalized applications.

Nowadays, the prevailing emphasis in survival analysis revolves around leveraging the representational power of deep neural networks to model intricate time-to-event relationships. Following this trend, \cite{katzman2018deepsurv, kvamme2019time} replaced the linear model in the standard CPH model with deep neural networks and empirically showed superior performance. Ren et al.~\cite{ren2019deep} harnessed the power of the deep recurrent neural network~\cite{hochreiter1997long} to model the conditional hazard rate and its cumulative distribution over time without making any specific assumption. Chapfuwa et al.~\cite{chapfuwa2018adversarial} proposed an adversarial strategy to model the non-parametric time-to-event distribution with a neural network. However, these deep neural networks are often regarded as "black boxes", which presents a challenge to uncover their decision-making processes.

Conversely, tree-based methods for survival analysis, such as survival tree~\cite{gordon1985tree} and random survival forest~\cite{ishwaran2008random}, provide a distinct advantage in terms of interpretability. These tree-based methods are typically built by employing a greedy expansion that enforces local homogeneity at each node split, potentially deviating from the global optimal solution. In addition, the traditional axis-aligned tree-splitting rule complicates their ability to capture complex interactions among covariates. Consequently, these tree-based survival models are empirically shown to be inferior to the deep-learning-based survival models in performance~\cite{katzman2018deepsurv, kvamme2019time, ren2019deep}.

Balancing interpretability and representational capacity is a typical trade-off in designing data-driven models. However, the neural parameterization of an oblique tree from a ReLU network~\cite{lee2020oblique} enables us to learn an interpretable deep neural network. 
In this paper, we propose a neural survival oblique tree (termed NSOTree) for survival analysis by leveraging the neural parameterization of the oblique tree. 
The contribution of this work is four-fold: 
\begin{itemize}
    \item  We propose a neural survival oblique tree (termed NSOTree) that harnesses the representational power of deep neural networks while maintaining interpretability.  
    \item To further enhance NSOTree's interpretability, we propose to introduce sparsity into its oblique split rule by employing a soft-thresholding strategy. 
    \item The NSOTree can be seamlessly integrated into statistical survival models in a plug-and-play fashion to harness their model-based advantages. We demonstrated the CPH model in this paper; while other statistical survival models can be alternatives. 
    \item  Our empirical findings showed that the proposed NSOTree showed superior performance compared to Cox-based deep survival models and tree-based survival methods.
\end{itemize}

\section{Related Work}
\label{sec:headings}
\subsection{Deep Learning in Survival Analysis}
In 1995, Faraggi and Simon~\cite{faraggi1995neural} initiated the endeavor to employ a deep neural network as a substitute for the linear model in standard CPH. 
Since its first attempt, deep neural networks have risen to prominence in data-driven survival analysis. With the advancement of deep learning in the past decade, the DeepSurv model~\cite{katzman2018deepsurv}, an extension of the Faraggi-Simon network that incorporates modern deep-learning techniques (e.g., activation function, optimizer, and regularization) has shown superior performance against the linear CPH model. To further improve the scalability and reduce the computational overhead of the DeepSurv model, Kvamme et al.~\cite{kvamme2019time} proposed a case-control approximation to the Cox's partial likelihood. This modification enables the CPH model to be optimized using modern deep learning techniques such as stochastic gradient descent. 
Apart from the deep CPH models, assumption-free deep survival models have also been explored by the previous work. Following this line of work, Lee et al.~\cite{lee2018deephit} proposed to directly learn the distribution of survival times through a deep neural network. To further capture the sequential patterns between adjacent time slots, Ren et al. took advantage of the deep recurrent network to model the conditional probability of the event. In parallel, Chapfuwa et al.~\cite{chapfuwa2018adversarial} proposed an adversarial learning scheme to model the non-parametric time-to-event distribution. 

Even though deep survival models have achieved state-of-the-art performance, their interpretability remains a challenge due to their "black-box" nature. To overcome this issue, we plugged the proposed NSOTree into the CPH to gain interpretability while harnessing the representational power of deep neural networks.


\subsection{Tree-based Methods in Survival Analysis}
Conventional tree-based survival methods are competing approaches to the deep survival models. One notable work is the survival tree that~\cite{gordon1985tree} retains the main structure of a decision tree while performing a Log-rank statistical test to govern node splitting. The random survival forests~\cite{kogalur2008random} extended the survival tree by leveraging ensemble learning. As opposed to assuming local homogeneity in the survival tree, 
Wang et al.~\cite{wang2016functional} proposed to maximize the heterogeneity between branches within the same split for time-to-event modeling. To further capture the interactions between covariates in survival analysis, Kretowska et al.~\cite{kretowska2019oblique} proposed to leverage a non-axis-aligned oblique split. Jaeger et al.~\cite{jaeger2019oblique} further proposed oblique random survival forests to harness the power of ensemble learning. 

Although tree-based methods are advantageous in interpretability, their expansion embraces a greedy strategy, assuring local optima at each node split. However, this greedy expansion can potentially be divergent from the global optimal solution, as it is not tailored for global optimization. On the contrary, the proposed NSOTree encourages the approximation to the global optima through end-to-end optimization. In addition, the proposed NSOTree has the potential to model complex time-to-event distribution because its neural parameterization and oblique split rule enable it to approximate arbitrarily complex functions (Details will be discussed in Section~\ref{sec:notree}, and supporting evidence can be found in Figure~\ref{fig:linear} and \ref{fig:gau}).

\section{Problem Formulation}
With the random variable $t$ representing the time of occurrence for the event of interest, the objective is to model the distribution of $t$. This is commonly formulated to compute the survival function $S(t)$ (i.e., the probability that an event occurs beyond time $t$) of an event time $T$:  
\begin{equation}
    S(t) = \mathbb{P}(T>t) = 1 - \int_{0}^{t} f(s) \ ds
\end{equation}
where $f(t)$ denotes the probability density function. Alternatively, the hazard rate:
\begin{equation}
    h(t) = \frac{f(t)}{S(t)} = \lim_{\Delta t \rightarrow 0} \frac{\mathbb{P}(t \leq T < t + \Delta t | T \geq t)}{\Delta t},
\end{equation}
which specifies the risk at time $t$, is rather commonly used in certain survival models (e.g., Cox propositional hazard model). It is worth noting that the survival function can be recovered once the hazard rate is known to us. This is achieved through exponentiating the negative cumulative hazard rate $H(t)=\int_{0}^{t}h(s) \ ds$ by 
\begin{equation}
\begin{split}
     S(t) &= \exp(-H(t)).
\end{split}
\end{equation}

In practice, the survival data typically exhibits to be right-censored due to the lack of a follow-up track. It is given in the form of $N$ triples $\{\boldsymbol{x}, T, E\}_{n=1}^{N}$, where $\boldsymbol{x}, T, E$ denote the $d-$dimensional feature vector of the event, the time of record, and an event indicator ($0$ for censored observation and $1$ otherwise), respectively. Due to the right censorship, we view the survival analysis as a semi-supervised learning task through a direct mapping from $\boldsymbol{x}$ to hazard rate $h(t)$. As a result, we considered the Cox Proportional Hazard (CPH) model as an example in the rest of this paper due to its popularity and success in survival analysis, while other statistical survival models can be alternatives.  

The CPH model links the feature vector $\boldsymbol{x}$ and hazard rate $h(t)$ in a semi-parametric fashion:
\begin{equation}
    h(t | \boldsymbol{x}) = h_0(t) \exp(g_{\theta}(\boldsymbol{x})), 
\end{equation}
where $\exp(g_{\theta}(\boldsymbol{x}))$ is the fully-parametric relative risk function $h_{\theta}(\boldsymbol{x})$ parameterized by $\theta$. The estimation of $\theta$ can achieved by maximizing the Cox partial likelihood (see Section~\ref{sec:partial} for details). $h_0(t)$ denotes the baseline hazard, which needs to be evaluated separately after the estimation of $\theta$.
In the standard form of CPH, $g_{\theta}(\cdot)$ is governed by a linear model:
\begin{equation}
    g_{\theta}(\boldsymbol{x}) = \theta^T \boldsymbol{x}.
\label{eqn:cph}
\end{equation}
Recently, this linear parameterization of $g_{\theta}(\cdot)$ has been replaced by more advanced models to harness their representational capacity, e.g., multi-layer perceptron~\cite{faraggi1995neural, katzman2018deepsurv, kvamme2019time}. In this paper, we leverage the neural oblique tree to propose a new parameterization of $g_{\theta}(\cdot)$. 


\section{Methodology}
\label{sec:nsot}

As opposed to the axis-aligned tree split, which involves splitting a single feature at each node, the oblique split rule considers all features simultaneously at each decision node through their linear combination.
\begin{Definition}
    The decision hyperplane of the oblique split is governed by 
\begin{equation}
    \boldsymbol{\alpha}_i^T \boldsymbol{x} \geq b_i, \textbf{x}\in \mathbb{R}^{d}
\label{eqn:oblique}
\end{equation}
where $\boldsymbol{\alpha}_i \in \mathbb{R}^{d}$ and $b_i \in \mathbb{R}^{1}$ are the parameters at the $i$-th splitting node.
\end{Definition}
 When $\boldsymbol{\alpha}_i$ has only a single entry of $1$, the oblique split rule reduces to an axis-aligned split condition. 

\subsection{Neural Survival Oblique Tree}
\label{sec:notree}
At first glance, non-parametric tree-based methods might appear to contrast with parametric deep neural networks. However, numerous studies~\cite{yang2018deep,lee2020oblique,kontschieder2015deep} have seamlessly integrated the two in an end-to-end fashion. One representative work is to leverage the local piece-wise constant in a neural network~\cite{lee2020oblique}. A typical deep neural network can be broken down into linear operations and non-linear activation. Without loss of generality, we take the multi-layer perceptron (MLP) as an example. Given a $L$-layer MLP, the output from the $l$-th layer with $l \in \{1, \cdots, L\}$  can be expressed as 
\begin{equation}
\begin{split}
     \boldsymbol{a}^{(l)} = \sigma(\boldsymbol{W}^{(l)} \boldsymbol{a}^{(l-1)} + \boldsymbol{b}^{(l)}), \
     \boldsymbol{a}^{(0)} = \textbf{x},
\end{split}
\end{equation}
where $\sigma(\cdot)$ denotes the activation function, and $\boldsymbol{W}^{(l)}$ and $\boldsymbol{b}^{(l)}$ are the weight matrix and bias vector, respectively. Intuitively, the linear transformation within each layer of an MLP aligns with the oblique split rule (Eqn.~\ref{eqn:oblique}) without considering the effect of the non-linear activation function. 

\begin{theorem}\cite{lee2020oblique}
For an MLP $f_{\theta}(\cdot)$ parameterized by $\theta$ with piece-wise linear activation function $\sigma(\cdot)$,  the output of the entire network is the affine transformation from all the hidden layers.
\label{theorem:1}
\end{theorem}
Theorem~\ref{theorem:1} immediately implies that the end-to-end behavior of the network $f_{\theta}$ is contingent upon the activation pattern (i.e., the activated neurons at each linear transformation within a layer). Take ReLU as an example of piece-wise linear activation function:
\begin{equation}
\begin{split}
     & [\sigma_{\text{ReLU}}(\boldsymbol{z}^{(l)})]_j = \max(\boldsymbol{z}^{(l)}_j, 0) = \boldsymbol{a}^{(l)}  \\ 
     & \ \ \ \text{with} \ \boldsymbol{z}^{(l)} = \boldsymbol{W}^{(l)} \boldsymbol{a}^{(l-1)} + \boldsymbol{b}^{(l)},
\end{split}
\label{eqn:relu}
\end{equation}
where $\max(\cdot)$ operates element-wise. The activation pattern of the ReLU activation, depicted by its derivative, is piece-wise constant. Mathematically,
\begin{equation}
    \boldsymbol{o}^{(l)}_j = \frac{\partial  \sigma_{\text{ReLU}}(\boldsymbol{z}_j^{(l)})}{\partial \boldsymbol{z}_j^{(l)}} = \mathbbm{1}[\boldsymbol{z}^{(l)}_j \geq 0],
\end{equation}
where operator $\boldsymbol{o}^{(l)}_j$ denotes derivative. 
In essence, the activation pattern of ReLU determines $\boldsymbol{z}^{(l)}_j \geq 0$.  
Accordingly, once the activation pattern is fixed (i.e. when the information $\boldsymbol{z}^{(l)}_j \geq 0$ is known), the non-linear ReLU activation function degenerates to a linear function as $\boldsymbol{z}^{(l)} \times \boldsymbol{o}^{(l)}$. 
It is obvious that any function taking $\boldsymbol{o}^{(l)}$ as input will remain piece-wise constant. Intuitively, this characteristic aligns with the concept of local homogeneity in tree construction.

\begin{figure*}[!t]
    \centering
    \includegraphics[width=1.0\textwidth]{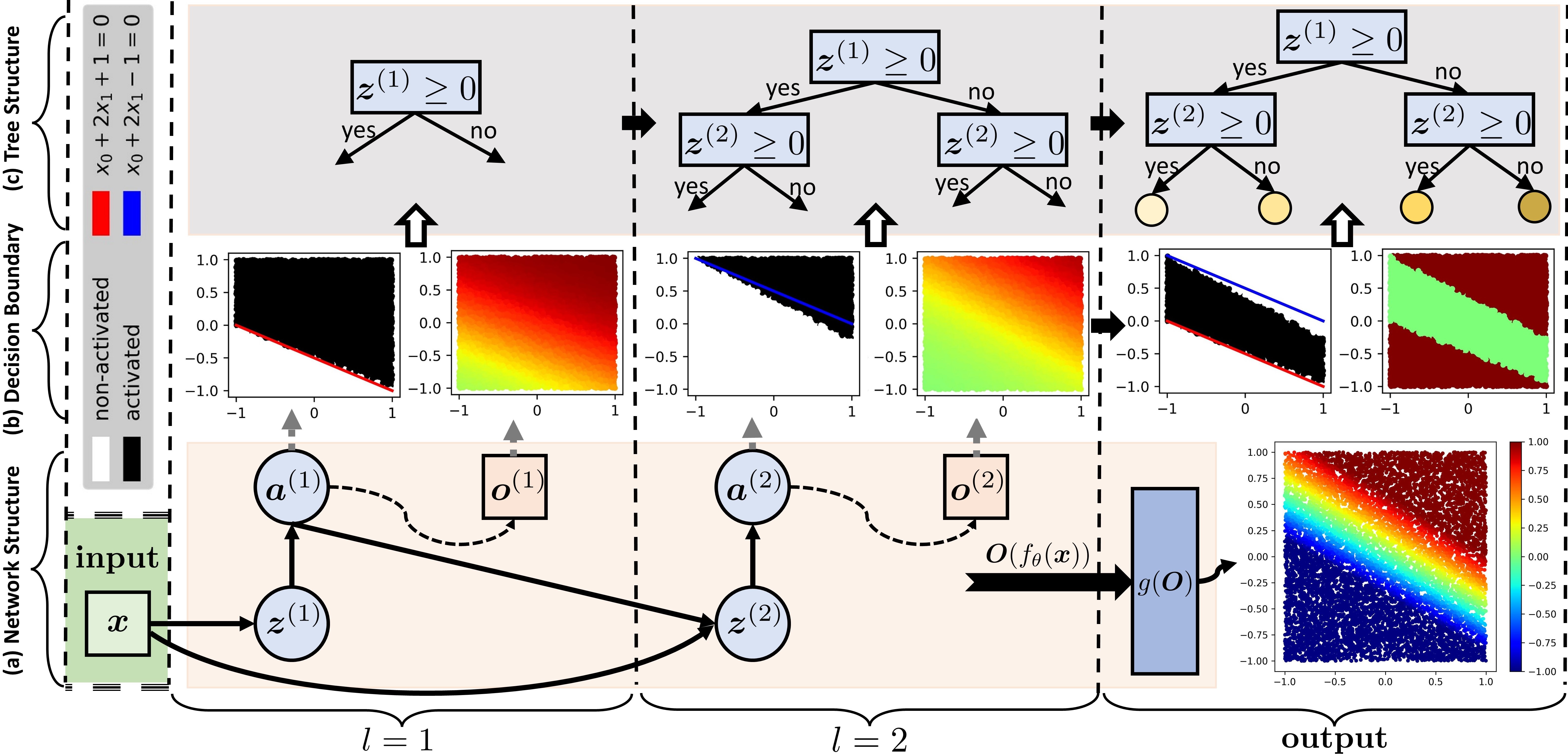}
    \caption{Illustrative example for constructing a neural survival oblique tree (NSOTree) to approximate a linear hazard function $h_{\theta}(\boldsymbol{x})=x_0 + 2 x_1$. Subpanel figures (a), (b), and (c) represent the network structure of the proposed NSOTree, the decision boundary at each oblique split, and the resulting tree representation, respectively.}
    \label{fig:workflow}
\end{figure*}

\noindent \textbf{Single-layer Neural Survival Oblique Tree:}
Formally, considering a $f_{\theta}$ with one layer, the oblique split in Eqn.~\ref{eqn:oblique} is equivalent with Eqn.~\ref{eqn:relu}. The leaf node (i.e., hazard rate $h_{\theta}(\boldsymbol{x})$) can be derived from the activation patterns through a direct mapping $g: \boldsymbol{o}^{(1)}\rightarrow h_{\theta}(\boldsymbol{x})$. Up to now, we constructed an oblique tree with a depth of one (see Figure~\ref{fig:workflow}(a) where $l=1$). 

\noindent \textbf{Multi-layer Neural Survival Oblique Tree:}
The construction of a single-layer neural oblique tree can be extended to multi-layer cases. This is because a multi-layer $f_{\theta}$ shares the same property as the single-layer one (see Theorem~\ref{theorem:1}). To do so, a specified structure of $f_{\theta}$ needs to be constructed. Unlike standard MLPs, a neural oblique survival tree's $l$-th layer requires taking $\boldsymbol{x}$ as an extra input for constructing oblique split rules (see Eqn.~\ref{eqn:oblique}), going beyond the reliance on the previous layer's output. Mathematically, the newly constructed neurons at the $l$-th node of a neural oblique survival tree can be formulated as 
\begin{equation}
    \boldsymbol{z}^{(l)} = \boldsymbol{W}^{(l)} \text{concat}[\boldsymbol{x} , \boldsymbol{a}^{(l-1)}] + \boldsymbol{b}^{(l)} \ \forall l > 2,
\end{equation}
where $\text{concate}[\cdot]$ denotes concatenation of vectors.
When $l=1$, this construction reduces to a single-layer case. It is important to highlight that the dimension of the weight matrix is $\boldsymbol{W}^{(l)} \in \mathbb{R}^{(d+l-1) \times d_{h}}$  where the default value for the hidden dimension of neurons $d_h$ is set to 1.
We further define the notation $\boldsymbol{O}(f_{\theta}(\boldsymbol{x}))$ as a collection of activation patterns ($\boldsymbol{O}=\{\boldsymbol{o}^{(1)}, \cdots, \boldsymbol{o}^{(L)}\}$) for notation brevity. Analogous to the single-layer case, the leaf output can be represented as $g: \boldsymbol{O}(f_{\theta}(\boldsymbol{x}))\rightarrow h_{\theta}(\boldsymbol{x})$. So far, we have constructed the multi-layer neural survival oblique tree (see Figure~\ref{fig:workflow}(a)). 

\begin{theorem}\cite{lee2020oblique}
    The derived neural oblique tree exhibits the same representational equivalence as standard oblique trees.
\label{theorem:2}
\end{theorem}
The proof of Theorem~\ref{theorem:2} can be trivial for a single-layer case. In the case of multiple layers, we provided a two-layer illustrative example in approximating a linear rate function $h_{\theta}(\boldsymbol{x})=x_0 + 2 x_1, \forall x_0 \in [-1, 1), x_1 \in [-1, 1)$ in this paper (Figure~\ref{fig:workflow}); while the complete proof is available in~\cite{lee2020oblique} (Theorem 1 and 2). In this example, we can see that the activation patterns of the first layer ($\boldsymbol{o}^{(1)}$) and second layer ($\boldsymbol{o}^{(2)}$) approximate the boundary of the target function from $x_0 + 2 x_1 + 1 = 0$ to $x_0 + 2 x_1 -1 = 0$ (see Figure~\ref{fig:workflow}(b)). The resulting activation patterns and structures constructed a two-layer oblique survival tree ((see Figure~\ref{fig:workflow}(c)). For simplicity, we considered the mapping function $g$ as another linear function in this example and the rest of the paper. Hence, we reuse the notation $g_{\theta}$ to denote the constructed neural oblique tree in the following derivation to keep consistency with Eqn.~\ref{eqn:cph}. 

The derived neural oblique survival tree can be optimized through back-propagation to approximate hazard rate function. However, in practice, the sub-differentiability of the ReLU function leads to an uninformative gradient w.r.t. model parameters $\theta$ when $\boldsymbol{o}_j^{(l)} = 0$. Equivalently, one branch of each split (i.e., whenever $\boldsymbol{z}^{(l)}_j < 0$) can not provide valuable information to learn the network through backpropagation. To mitigate this challenge, the ReLU function is replaced with the Softplus function in our experiment, which is a smooth approximation to the ReLU function:
\begin{equation}
     [\sigma_{\text{Softplus}}(\boldsymbol{z}^{(l)})]_j = \log(1 + \exp(z_j^{(l)})).
\label{eqn:softplus}
\end{equation}
It is worth noting that we abandoned the weighted combination of ReLU and Softplus as outlined in~\cite{lee2020oblique}:
\begin{equation}
    [\sigma(\boldsymbol{z}^{(l)})]_j = \lambda_t  [\sigma_{\text{ReLU}}(\boldsymbol{z}^{(l)})]_j + (1 - \lambda_t) [\sigma_{\text{Softplus}}(\boldsymbol{z}^{(l)})]_j,
\end{equation}
where $\lambda_t$ is annealed at each iteration during training. Instead, we empirically discovered that the Softplus function solely demonstrated superior performance.

\subsection{NSOTree in Survival Analysis}
\label{sec:partial}
The derived NSOTree can be explicitly plugged into the CPH model due to its end-to-end parameterization. Therefore, its optimization can be achieved by maximizing the Cox partial likelihood:
\begin{equation}
    \mathcal{L}_{\text{cox}} = \prod_{n: E_n =1} \left(  \frac{\exp(g_{\theta} (\boldsymbol{x}_n))}{\sum_{k\in \mathcal{R}(T_n)} \exp(g_{\theta} (\boldsymbol{x}_k) ) } \right),
\end{equation}
where $\mathcal{R}(T_n)$ is the risk set (i.e., set of all samples that are at risk at time $T_n$). The corresponding negative log-likelihood loss is given by
\begin{equation}
\begin{split}
    \mathcal{L}_{\text{cox-nll}} = - \sum_{n: E_n =1} \left(  g_{\theta} (\boldsymbol{x}_n)  - \log \sum_{k\in \mathcal{R}(T_n)}\exp(g_{\theta} (\boldsymbol{x}_k) )   \right).
\end{split}
\end{equation}
In practice, we optimize the Cox negative log-likelihood loss in~\cite{kvamme2019time} to take advantage of the mini-batch stochastic gradient descent in modern neural network optimization: 
\begin{equation}
    \mathcal{L}_{\text{cox-nll}} = \sum_{n: E_n =1} \log \left( \sum_{k\in \widetilde{\mathcal{R}}(T_n)}\exp(g_{\theta} (\boldsymbol{x}_k) - g_{\theta} (\boldsymbol{x}_n) )   \right),
\end{equation}
where $\widetilde{\mathcal{R}}(T_n)$ is a sufficiently large subset of the entire risk set, which implies a large batch size. 

Although we have focused on combining the NOTree with the CPH model in this paper, it is important to note that this approach can be generalized to other model-based forms of survival analysis, e.g., the Accelerated Failure Time (AFT) model~\cite{wei1992accelerated}. This augmentation can enhance both the model's representational capacity and its interpretability. The proposed approach can even be adopted in the model-free formulation~\cite{ranganath2016deep, chapfuwa2018adversarial} of survival analysis to take advantage of their model-specific advancement while improving interpretability. 


\subsection{Sparsified NSOTree}
Although the oblique split rule is advantageous in capturing more complex relationships (e.g., interactions between features) in the data than the axis-aligned rule, it poses challenges in interpreting the resulting decision boundaries. Specifically, the oblique split convolves multiple input features adding complexity to disentangle the importance of each feature. However, we seek to retain the interpretability inherent to tree-based methods when applied to survival analysis. To this end, we propose to sparsify the weight in the oblique split rule of the NSOTree. Considering the standard definition of the oblique split rule (Eqn.~\ref{eqn:oblique}), this is typically formulated as enforcing the $\mathcal{L}_1$ constraint to the weight $\boldsymbol{\alpha_i}$:
\begin{equation}
\begin{split}
    \min_{\boldsymbol{\alpha}_i} \ ||\boldsymbol{\alpha}_i||_1 \ \text{s.t.} \  \boldsymbol{\alpha}_i^T \boldsymbol{x} \geq b_i. 
\end{split}
\end{equation}
However, the sub-differentiable $\mathcal{L}_1$ norm raises issues in the optimization phase, especially for greedy expanded oblique trees. Even in the case of  NSOTree, which can be optimized end-to-end, we empirically revealed that directly adding $\mathcal{L}_1$ penalty to the loss function led to training instability. Alternatively, we leverage the soft-thresholding operator~\cite{tibshirani1996regression} to approximate the $\mathcal{L}_1$ sparsity regularization. The soft-thresholding operator can be derived from the proximal operator of $\mathcal{L}_1$ norm:
\begin{equation}
    [\text{prox}_f(\boldsymbol{\alpha}_i)]_j = \text{sign}([\boldsymbol{\alpha}_i]_j)\max([|\boldsymbol{\alpha}_i|]_j - \lambda, 0), 
\end{equation}
where $\lambda > 0$ controls the regularization strength. 

In the context of NSOTree, the $\mathcal{L}_1$ regularization can be achieved by recursively soft-thresholding the weight matrix $\boldsymbol{W}^{(l)}$ at each layer:
\begin{equation}
    \Tilde{\boldsymbol{W}}^{(l)} = [\text{prox}_f(\boldsymbol{W}^{(l)})]_j, \ \forall l \in \{1, \cdots, L\}.
\end{equation}





\section{Experiments and Results}
We conducted experiments on simulated and real survival data and additionally compared the proposed method to strong baselines in order to thoroughly validate its effectiveness.

\subsection{Dataset}
We conducted experiments on two simulated data to check the sanity of the proposed model and gain insight into its representational capacity.
The simulated datasets were sourced from~\cite{katzman2018deepsurv},  consisting of a training ($N = 4000$), validation ($N = 1000$), and testing ($N = 1000$) set. Each sample includes $d = 10$ covariates drawn from a uniform distribution on $[-1,1)$. The event time $T$ was generated based on an exponential Cox model.  Further information about the simulation process can be found in~\cite{katzman2018deepsurv}. These simulated datasets encompass two subcategories characterized by either a linear risk function or a Gaussian risk function.

\noindent \textbf{Linear Risk Function:}
The risk function is a linear function w.r.t. the first two covariates $x_0$ and $x_1$: 
\begin{equation}
h(x) = x_0 + 2x_1
\end{equation}

\noindent \textbf{Gaussian Risk Function:}
The risk function was formulated as a Gaussian function w.r.t. the first two covariates $x_0$ and $x_1$: 
\begin{equation}
h(\boldsymbol{x}) = \log(\lambda_{max}) \exp(-\frac{x_0^2+x_1^2}{2r^2}),
\end{equation}
where $\lambda_{max}=5.0$ and a scale factor of $r = 0.5$.

Beyond the two simulated datasets, we carried out additional experiments on two real survival datasets for which both the true risk function and the true event time distribution are unknown.

\noindent \textbf{SUPPORT:}
The SUPPORT dataset, originating from the Study to Understand Prognoses Preference Outcomes and Risks of Treatment~\cite{knaus1995support}, provides detailed information on the survival duration of critically ill hospitalized adults. It comprises 8873 patients (patients with missing features were dropped) with 14 covariates.  The censorship rate within this dataset stands at approximately $32\%$.

\noindent \textbf{METABRIC:}
The Molecular Taxonomy of Breast Cancer International Consortium (METABRIC) dataset is designed to improve breast cancer treatment recommendations~\cite{curtis2012genomic}. 
The dataset includes gene expression and clinical data for 1,980 patients. Of these, $57.72\%$ experienced breast cancer-related deaths with a median survival of 116 months. The data was processed using the IHC4+C test, integrating four gene indicators (MKI67, EGFR, PGR, and ERBB2) with clinical features, yielding a total of 9 covariates. The proposition of censorship is around $42\%$.

\subsection{Evaluation metrics}
The evaluation of the performance for survival analysis was carried out by the time-dependent concordance index (\textbf{C-index}) and the \textbf{Brier Score}. Specifically, the C-index is a generalization of the area under the ROC curve (AUC), designed to assess a model's ability to predict the ordering of event times through a ranking mechanism. Similar to the AUC, the C-Index attains a value of 1 for a perfect ranking of event time and 0.5 for a random guess. The Brier Score measures the mean squared difference between predicted probabilities and the actual outcomes of events. In essence, the Brier Score indicates whether an event of interest occurs before or after a given time $t$.  We use the integrated Brier Score (IBS) to compute the Brier Score over an interval instead of a fixed time.

\subsection{Baselines}
\begin{itemize}
    \item \textbf{CPH-Linear} is Cox proportional hazard model, where $g_\theta(\cdot)$ is a linear model. 
    \item \textbf{CPH-MLP (DeepSurv)} is a neural parameterized CPH model, termed DeepSurv, where $g_\theta(\cdot)$ is parameterized by a multi-layer perceptron (MLP). 
     \item \textbf{CPH-MLP (CC)} is a variant of DeepSurv that relaxes the proportionality constraint in the CPH model. The resulting case-control approximation of Cox partial likelihood made it compatible with stochastic gradient descent.
    \item \textbf{ST} is the survival tree that maximizes the survival probabilities between branches at each split by leveraging Log-rank statistics. 
    \item \textbf{RSF} is the random survival forest that ensembles multiple survival trees using bootstrapped samples. 
\end{itemize}

\subsection{Experiment Setup and Implementation}
In our experiments with the simulated datasets, we adhered to the default training, validation, and testing data splits for training and evaluating all the methods. All the covariates were standardized to have a zero mean and unit variance. The 95\% confidence intervals were obtained by bootstrapping the testing dataset and sampled the 2.5\% and 97.5\% percentile of the resulting C-index, as outlined in~\cite{katzman2018deepsurv}. 

We conducted 5-fold cross-validation for the real datasets (i.e., SUPPORT and METABRIC) to thoroughly assess the performance and generalizability of all methods. The mean and standard deviation of the C-index are reported (mean $\pm$ std) in Table~\ref{tab:all_results}. For feature engineering, we applied one-hot encoding to all categorical variables and standardized all numerical variables to have a zero mean and unit variance. 

The proposed model was built on top of the \textbf{pycox}\footnote[1]{pycox python package: \url{https://github.com/havakv/pycox}} python package. 
The hyperparameter tuning for the proposed model was carried out by the random search~\cite{bergstra2012random}. The optimal learning rating was found to be $0.1$. The batch size was set to 1024. We also implemented the CPH-MLP (DeepSurv) and CPH-MLP (CC) using the pycox python package. Whereas, the implementation of the CPH-Linear, ST, and RSF was adopted from the \textbf{scikit-survival}\footnote[2]{scikit-survival python package: \url{https://github.com/sebp/scikit-survival}} python package. The hyperparameter configurations of all baselines were adopted from~\cite{katzman2018deepsurv,kvamme2019time}. 
The code and all pretrained models will be made publicly available on GitHub: \href{https://github.com/xs018/NSOTree}{https://github.com/xs018/NSOTree}.

\begin{table*}[!t]
  \caption{Performance comparison on \textbf{C-index}. We reported the $95\%$ confidence interval of the C-index for the simulated dataset. Whereas, for the SUPPORT and METABRIC datasets, 5-fold cross-validation was conducted, and the mean ($\pm$ a standard deviation) of the C-index was reported. The best metrics in each column are highlighted in bold.}
  \centering
  \resizebox{\textwidth}{!}{
  \begin{tabular}{l|c|c|c|c}
    \toprule
    Methods  & Simulated Linear & Simulated Gaussian  &  SUPPORT & METABRIC  \\
    \hline
     CPH-Linear & 0.7737 (0.772, 0.775) & 0.5070 (0.505, 0.509) & 0.5689 $\pm$ 0.0099 &   0.6356 $\pm$ 0.0122  \\
    Cox-MLP (DeepSurv) & 0.7740 (0.772, 0.776) & 0.6489 (0.647, 0.651) & 0.6114 $\pm$ 0.0073  &  0.6517 $\pm$ 0.0122   \\
    Cox-MLP (CC) & 0.7753 (0.757, 0.789) & 0.6492 (0.627, 0.672) & 0.6130 $\pm$ 0.0029 &  0.6493 $\pm$ 0.0088   \\
    \bottomrule
     \toprule
    ST & 0.6940 (0.676, 0.715) & 0.5712 (0.544, 0.591)  &  0.5536 $\pm$ 0.0065  & 0.5632 $\pm$ 0.0144  \\
     RSF & 0.7650 (0.763, 0.766) & 0.6455 (0.643, 0.648) & 0.6140 $\pm$ 0.0043  &  0.6431 $\pm$ 0.0111  \\
    \textbf{NSOTree (Ours)} & \textbf{0.7797 (0.768, 0.795)} & \textbf{0.6595 (0.636, 0.678)}  & \textbf{0.6199 $\pm$ 0.0055} &  \textbf{0.6594 $\pm$ 0.0097}   \\
    \bottomrule
  \end{tabular}
  }
  \label{tab:all_results}
\end{table*}

\section{Results and Analysis}
\subsection{Representational Capacity in Modeling Event Rate}
The proposed NSOTree showed superior performance in estimating the event rate over time on both simulated datasets and real datasets in terms of the C-index (Table~\ref{tab:all_results}). Additionally, it showcased an exceptional representational capacity.

\begin{figure}[!b]
    \centering
    \includegraphics[width=\textwidth]{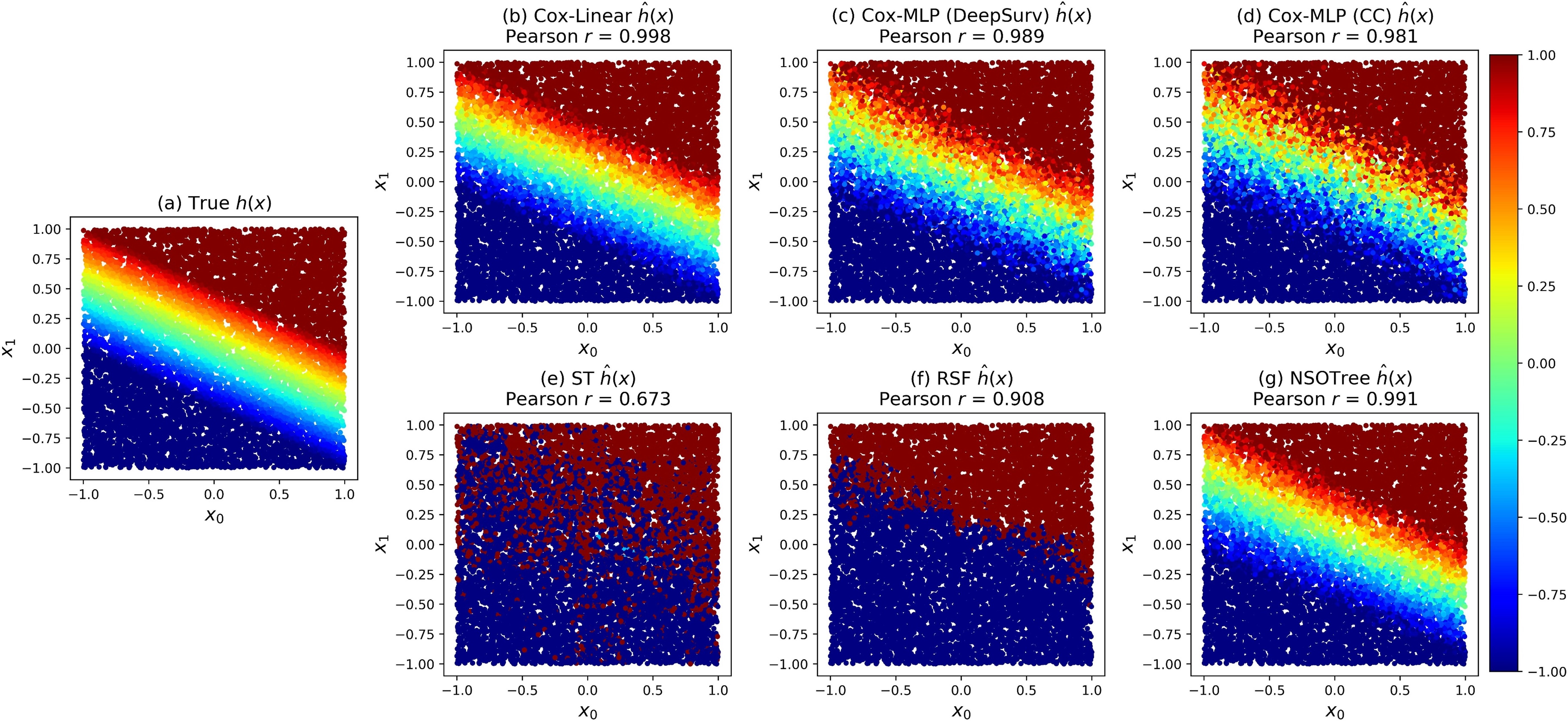}
    \caption{The predicted risk function on the simulated dataset versus the true Linear risk function.  The methods in subpanel figures are the same as in Table~\ref{tab:all_results}. The Pearson correlation coefficient ($r$) between the true risk function and the predicted ones is reported.}
    \label{fig:linear}
\end{figure}

\noindent \textbf{Result in Simulated datasets.} 
In our analysis of the simulated dataset with a linear risk function, we observed that all the Cox-based baselines and the proposed NSOTree exhibited similar performance (Table~\ref{tab:all_results}) and could effectively approximate the linear risk function (Figure~\ref{fig:linear} (b), (c), (d), and (g)). This can be attributed to the linear risk function's inherent simplicity. Therefore, introducing additional model complexity did not yield significant improvements. In contrast, the axis-aligned tree-based methods showed inferior performance even in modeling the linear risk function which is not parallel to the covariate's axis. More concrete evidence can be found in their predicted hazard function where we observed a significant deviation from the true hazard function (Figure~\ref{fig:linear} (e) and (f)).  

\begin{figure}[!t]
    \centering
    \includegraphics[width=\textwidth]{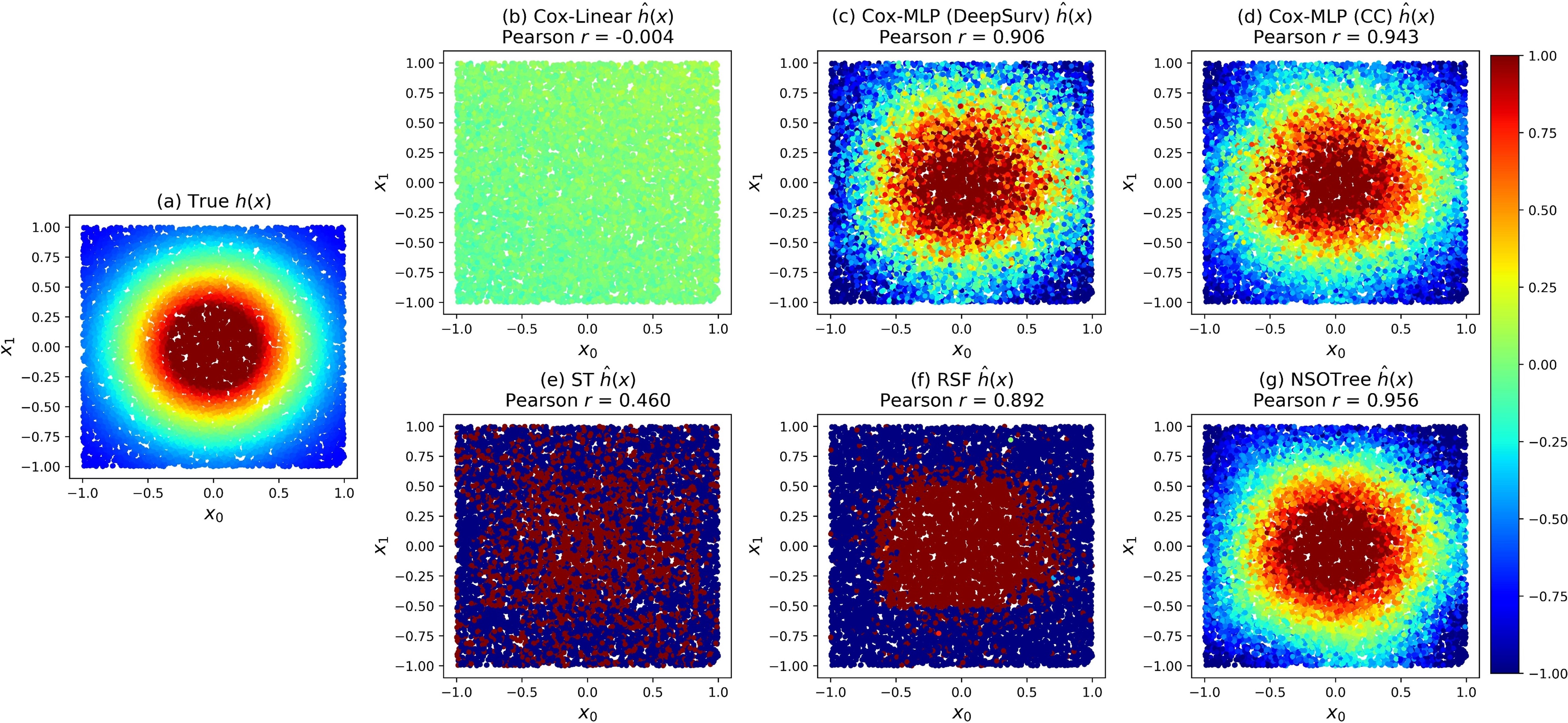}
    \caption{The predicted risk function on the simulated dataset versus the true Gaussian risk function.}
    \label{fig:gau}
\end{figure}

In the case of the Gaussian risk function, a similar phenomenon was observed (Table~\ref{tab:all_results} and Figure~\ref{fig:gau}). However, two noteworthy exceptions were observed. First, the proposed NSOTree showed obvious improvement over other Cox-based baselines with the best approximation (Pearson Correlation $r=0.956$) to the Gaussian risk function (see Figure~\ref{fig:gau} (g)). Second, the Cox-Linear model showed the worst performance due to the violation of the linear relationship between the risk function and covariates (see Figure~\ref{fig:gau} (b)).   

\noindent \textbf{Result in Real datasets.} 
In real datasets (see Table~\ref{tab:all_results}), we observed that the axis-aligned survival tree consistently showed worse performance compared to the deep learning-based methods.  Remarkably, the proposed NSOTree outperformed all the other deep learning-based baselines. Furthermore, in real-world scenarios, the NSOTree model we introduced beat the ensemble learning-based RSF method, even without the use of an ensemble. All observations supported that the proposed NSOTree gained exceptional representational capacity from the neural parameterization of the survival tree. 



\begin{figure*}[!t]
     \centering
    \includegraphics[width=\textwidth]{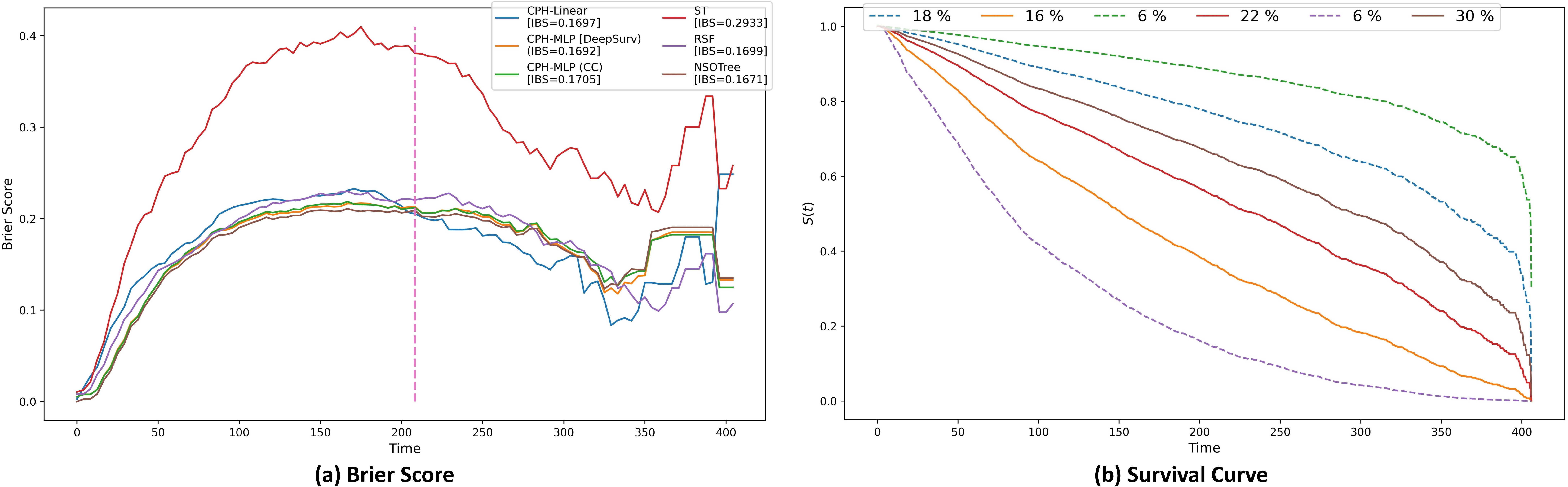}
    \caption{Visualization on the  METABRIC dataset: (a) Brier score produced by the baseline methods and proposed NSOTree. The legends are the corresponding IBS. (b) Survival curve cluster centers were generated by the proposed  NSOTree for the METABRIC dataset, which was carried out using K-means clustering (with 6 clusters). The legend provides details on the distribution of examples assigned to each cluster.}
    \label{fig:bs_surv}
\end{figure*}

\subsection{Event Time Prediction and Survival Curve}
We visualized the Brier Score (with IBS reported) and survival curve produced by the proposed method on the METABRIC dataset (Figure~\ref{fig:bs_surv}). We observed that the proposed NSOTree showed the lowest IBS compared to all baselines (Figure~\ref{fig:bs_surv} (a)), particularly for shorter durations ($t < 200$). Nonetheless, for longer time durations ($t > 300$), the performance of the proposed method is similar to COX-MLP (DeepSurv) and Cox-MLP (CC), but worse than CPH-Linear and RSF. In addition, the survival curve (Figure~\ref{fig:bs_surv} (b)) provided valuable information in predicting the survival time for the majority of the samples, except for the largest subgroup ($22 \%$). We also observed that there is a sudden drop in survival rates toward the end of the time duration (around $t=390$). 

\begin{figure}[!t]
    \centering
    \includegraphics[width=1.0\textwidth]{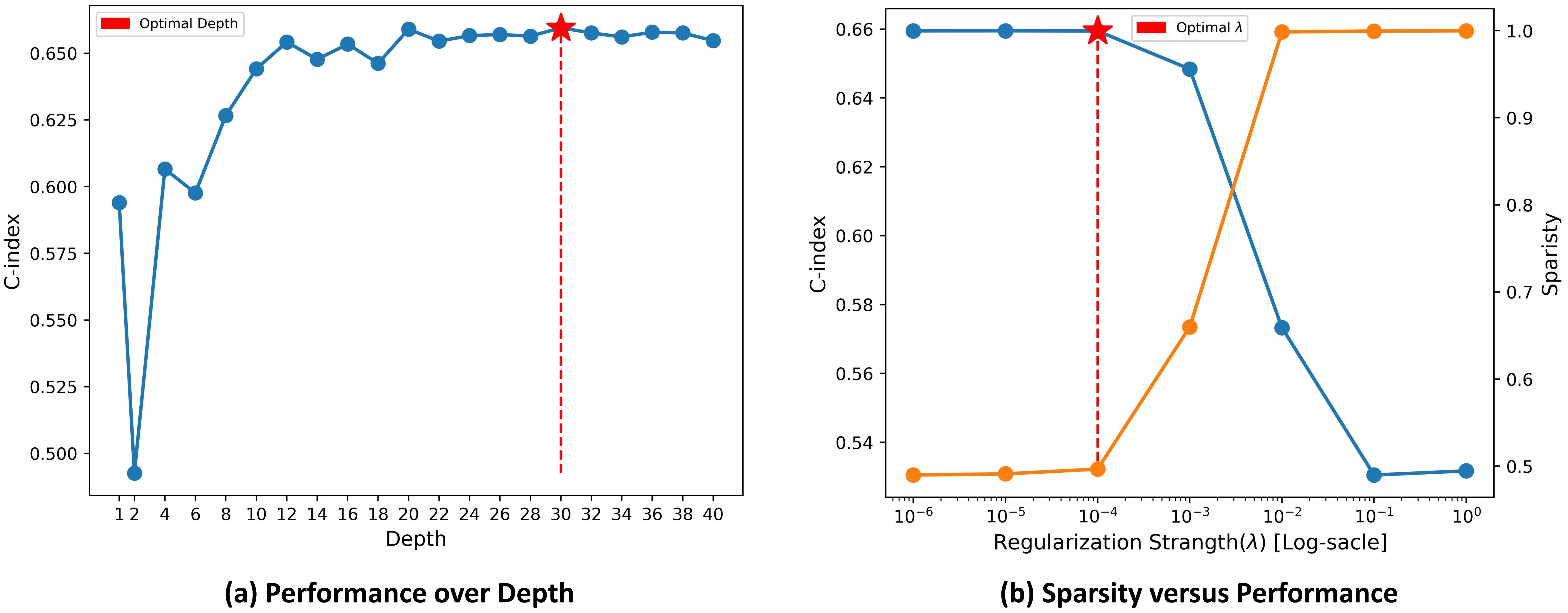}
    \caption{Ablation studies on depth and sparsity of the proposed NSOTree on the simulated dataset with Gaussian risk function: (a) performance over depth, and (b) Sparsity versus performance of the NSOTree with a fixed depth of 30.}
    \label{fig:ablation}
\end{figure}

\begin{figure}[!t]
    \centering
    \includegraphics[width=1.0\textwidth]{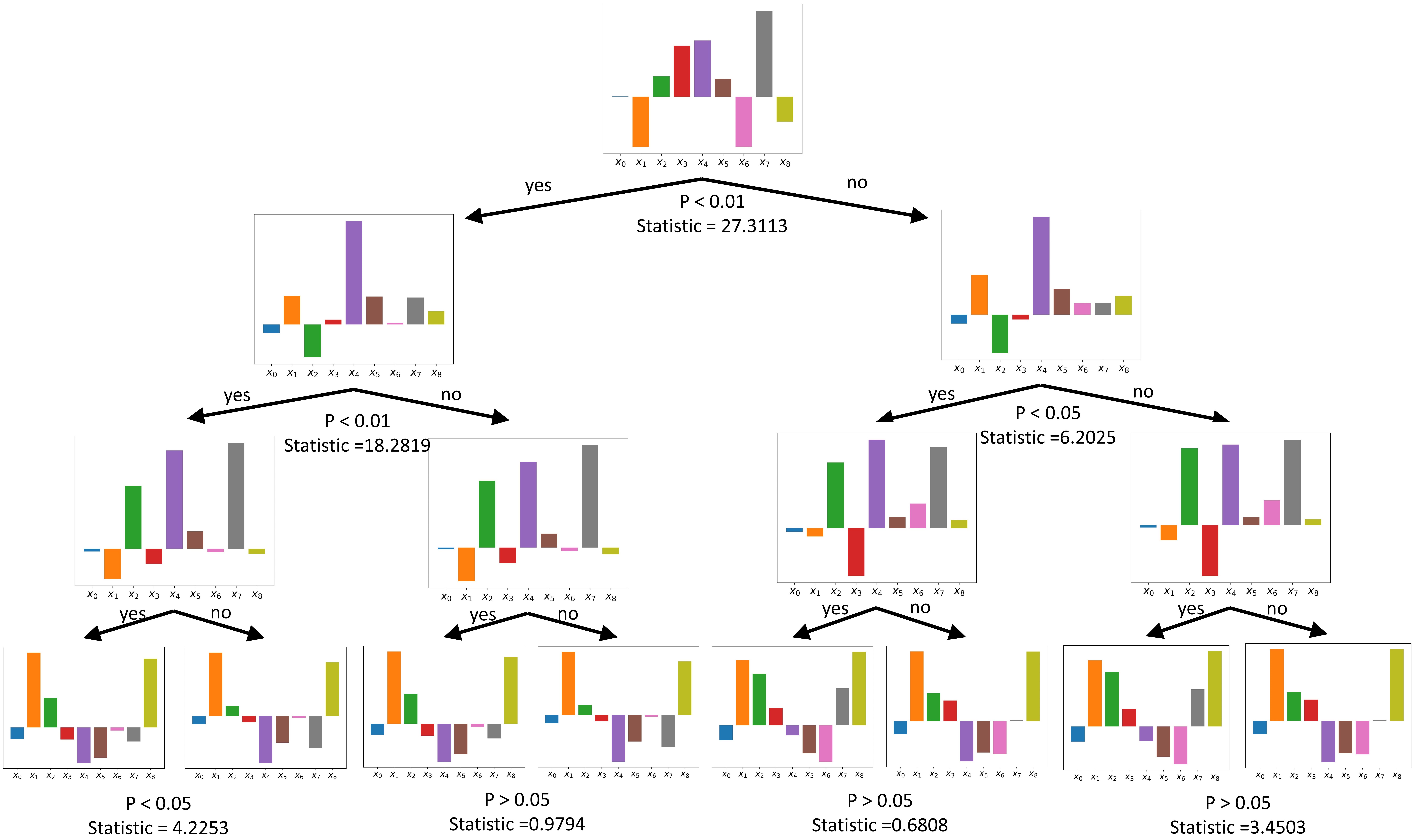}
    \caption{Visualization of the learned NSOTree with its first three layers on the METABRIC dataset where each bar plot denotes the weights at each oblique split. The p-values and statistics obtained from the Log-rank test between branches under the same split are reported.}
    \label{fig:tree}
\end{figure}

\subsection{Ablation studies}
The depth of trees is commonly a hyperparameter to tune, which is also the case for the proposed NSOTree. Additionally, the regularization strength, which controls the sparsity of the oblique split rule's weight, can also be adjusted to strike a balance between interpretability and performance. Thus, we conducted ablation studies on depth and sparsity.

\noindent \textbf{Performance over Depth:}
Upon varying the tree depth from 1 to 40 with an increment of 2, a common observation was that increasing the depth of the NSOTree generally led to enhanced performance (see Figure~\ref{fig:ablation} (a)). However, we also noted that beyond a certain threshold (e.g., $L > 20$), further increasing the tree depth did not substantially improve performance.

\noindent \textbf{Sparsity  vs. Performance:}
Our observations indicated a general trade-off between sparsity and performance (see Figure~\ref{fig:ablation} (b)). The NSOTree achieved its highest performance when $\lambda = 1\times10^{-6}$. Nevertheless, we opted for $\lambda = 1\times10^{-4}$ as the optimal choice, as it struck a balanced equilibrium between achieving sparsity and maintaining performance. This could potentially enhance interpretability in practical scenarios.

\subsection{Interpretability}
We visualized the first three layers of the constructed NSOTree using the METABRIC dataset (see Figure~\ref{fig:tree}). The constructed NSOTree shares the same structure as the standard oblique tree, where its decision-making processes can be easily retrieved. The NSOTree also provided valuable partitions of survival data based on the Log-rank statistics st each splitting node (see Figure~\ref{fig:tree}), which can benefit real-world applications. An example can be found in modern precision medicine where this partition helps to make personalized treatment plans for each subgroup. 

\section{Conclusion}
In this paper, we propose a Neural Survival Oblique Tree (NSOTree) for modeling time-to-event data. This model leverages the strengths of the tree structure, offering enhanced interpretability for "black-box" neural networks. It has the potential to uncover intricate relationships between covariates aiding in making informed decisions. Furthermore, it mitigates the inherent constraints of traditional greedy tree expansion, notably their inability to attain global optima. Experimental results from simulated datasets and real clinical datasets marked the superiority of the proposed NSOTree in comparison to other competitive survival models. 



\bibliographystyle{siamplain}
\bibliography{refs}

\begin{thebibliography}{10}

\bibitem{lee2003statistical}
{\sc E.~T. Lee and J.~Wang}, {\em Statistical methods for survival data
  analysis}, vol.~476, John Wiley \& Sons, 2003.

\bibitem{kaplan1958nonparametric}
{\sc E.~L. Kaplan and P.~Meier}, {\em Nonparametric estimation from incomplete
  observations}, Journal of the American statistical association, 53 (1958),
  pp.~457--481.

\bibitem{cox1992regression}
{\sc D.~R. Cox}, {\em Regression models and life-tables. breakthroughs in
  statistics}, Stat. Soc, 372 (1992), pp.~527--541.

\bibitem{katzman2018deepsurv}
{\sc J.~L. Katzman, U.~Shaham, A.~Cloninger, J.~Bates, T.~Jiang, and
  Y.~Kluger}, {\em Deepsurv: personalized treatment recommender system using a
  cox proportional hazards deep neural network}, BMC medical research
  methodology, 18 (2018), pp.~1--12.

\bibitem{kvamme2019time}
{\sc H.~Kvamme, {\O}.~Borgan, and I.~Scheel}, {\em Time-to-event prediction
  with neural networks and cox regression}, arXiv preprint arXiv:1907.00825,
  (2019).

\bibitem{ren2019deep}
{\sc K.~Ren, J.~Qin, L.~Zheng, Z.~Yang, W.~Zhang, L.~Qiu, and Y.~Yu}, {\em Deep
  recurrent survival analysis}, in AAAI, vol.~33, 2019, pp.~4798--4805.

\bibitem{hochreiter1997long}
{\sc S.~Hochreiter and J.~Schmidhuber}, {\em Long short-term memory}, Neural
  computation, 9 (1997), pp.~1735--1780.

\bibitem{chapfuwa2018adversarial}
{\sc P.~Chapfuwa, C.~Tao, C.~Li, C.~Page, B.~Goldstein, L.~C. Duke, and
  R.~Henao}, {\em Adversarial time-to-event modeling}, in International
  Conference on Machine Learning, PMLR, 2018, pp.~735--744.

\bibitem{gordon1985tree}
{\sc L.~Gordon and R.~A. Olshen}, {\em Tree-structured survival analysis.},
  Cancer treatment reports, 69 (1985), pp.~1065--1069.

\bibitem{ishwaran2008random}
{\sc H.~Ishwaran, U.~B. Kogalur, E.~H. Blackstone, and M.~S. Lauer}, {\em
  Random survival forests},  (2008).

\bibitem{lee2020oblique}
{\sc G.-H. Lee and T.~S. Jaakkola}, {\em Oblique decision trees from
  derivatives of relu networks}, in ICLR, 2020.

\bibitem{faraggi1995neural}
{\sc D.~Faraggi and R.~Simon}, {\em A neural network model for survival data},
  Statistics in medicine, 14 (1995), pp.~73--82.

\bibitem{lee2018deephit}
{\sc C.~Lee, W.~Zame, J.~Yoon, and M.~Van Der~Schaar}, {\em Deephit: A deep
  learning approach to survival analysis with competing risks}, in AAAI,
  vol.~32, 2018.

\bibitem{kogalur2008random}
{\sc U.~Kogalur, H.~Ishwaran, E.~Blackstone, and M.~Lauer}, {\em Random
  survival forests}, Annals of Applied Statistics, 2 (2008).

\bibitem{wang2016functional}
{\sc Y.~Wang, K.~Ren, W.~Zhang, J.~Wang, and Y.~Yu}, {\em Functional bid
  landscape forecasting for display advertising}, in Machine Learning and
  Knowledge Discovery in Databases: European Conference, ECML PKDD 2016, Riva
  del Garda, Italy, September 19-23, 2016, Proceedings, Part I 16, Springer,
  2016, pp.~115--131.

\bibitem{kretowska2019oblique}
{\sc M.~Kretowska}, {\em Oblique survival trees in discrete event time
  analysis}, IEEE Journal of Biomedical and Health Informatics, 24 (2019),
  pp.~247--258.

\bibitem{jaeger2019oblique}
{\sc B.~C. Jaeger, D.~L. Long, D.~M. Long, M.~Sims, J.~M. Szychowski, Y.-I.
  Min, L.~A. Mcclure, G.~Howard, and N.~Simon}, {\em Oblique random survival
  forests}, The annals of applied statistics, 13 (2019), p.~1847.

\bibitem{yang2018deep}
{\sc Y.~Yang, I.~G. Morillo, and T.~M. Hospedales}, {\em Deep neural decision
  trees}, arXiv preprint arXiv:1806.06988,  (2018).

\bibitem{kontschieder2015deep}
{\sc P.~Kontschieder, M.~Fiterau, A.~Criminisi, and S.~R. Bulo}, {\em Deep
  neural decision forests}, in Proceedings of the IEEE international conference
  on computer vision, 2015, pp.~1467--1475.

\bibitem{wei1992accelerated}
{\sc L.-J. Wei}, {\em The accelerated failure time model: a useful alternative
  to the cox regression model in survival analysis}, Statistics in medicine, 11
  (1992), pp.~1871--1879.

\bibitem{ranganath2016deep}
{\sc R.~Ranganath, A.~Perotte, N.~Elhadad, and D.~Blei}, {\em Deep survival
  analysis}, in Machine Learning for Healthcare Conference, PMLR, 2016,
  pp.~101--114.

\bibitem{tibshirani1996regression}
{\sc R.~Tibshirani}, {\em Regression shrinkage and selection via the lasso},
  Journal of the Royal Statistical Society Series B: Statistical Methodology,
  58 (1996), pp.~267--288.

\bibitem{knaus1995support}
{\sc W.~A. Knaus, F.~E. Harrell, J.~Lynn, L.~Goldman, R.~S. Phillips, A.~F.
  Connors, N.~V. Dawson, W.~J. Fulkerson, R.~M. Califf, N.~Desbiens, et~al.},
  {\em The support prognostic model: Objective estimates of survival for
  seriously ill hospitalized adults}, Annals of internal medicine, 122 (1995),
  pp.~191--203.

\bibitem{curtis2012genomic}
{\sc C.~Curtis, S.~P. Shah, S.-F. Chin, G.~Turashvili, O.~M. Rueda, M.~J.
  Dunning, D.~Speed, A.~G. Lynch, S.~Samarajiwa, Y.~Yuan, et~al.}, {\em The
  genomic and transcriptomic architecture of 2,000 breast tumours reveals novel
  subgroups}, Nature, 486 (2012), pp.~346--352.

\bibitem{bergstra2012random}
{\sc J.~Bergstra and Y.~Bengio}, {\em Random search for hyper-parameter
  optimization.}, Journal of machine learning research, 13 (2012).

\end{thebibliography}

\end{document}